\pdfoutput=1
\documentclass[11pt]{article}

\usepackage{dcolumn}
\usepackage{xspace}
\usepackage{amsfonts}
\usepackage{amsmath}
\usepackage{graphicx}
\usepackage{booktabs}
\usepackage{multicol}
\usepackage{multirow}
\usepackage{makecell}
\newcolumntype{d}[1]{D{.}{.}{#1}}

\newcommand{\tabincell}[2]{\begin{tabular}{@{}#1@{}}#2\end{tabular}}

\newcommand{\eat}[1]{}
\newcommand{\paratitle}[1]{\vspace{1ex}\noindent \textbf{#1}}

\let\oldhat\hat
\renewcommand{\vec}[1]{\mathbf{#1}}
\renewcommand{\hat}[1]{\oldhat{\mathbf{#1}}}
\renewcommand{\matrix}[1]{\mathbf{#1}}

\newcommand{\eg}{\emph{e.g.,}\xspace}

\newcommand{\ie}{\emph{i.e.,}\xspace}

\usepackage[]{emnlp2021}

\usepackage{times}
\usepackage{latexsym}

\usepackage[T1]{fontenc}

\usepackage[utf8]{inputenc}

\usepackage{microtype}

\title{Context-aware Entity Typing in Knowledge Graphs}

\author{Weiran Pan$^{1}$, Wei Wei$^{1}$\thanks{\hspace{0.15cm}Corresponding author: Wei Wei.} \and Xian-Ling Mao$^2$\\
   $^1$Cognitive Computing and Intelligent Information Processing Laboratory, School of Computer \\ Science and Technology, Huazhong University of Science and Technology \\
   $^2$Department of Computer Science and Technology, Beijing Institute
   of Technology \\
  \texttt{panwr789@gmail.com, weiw@hust.edu.cn, maoxl@bit.edu.cn} \\
}

\begin{document}
\maketitle
\begin{abstract}
Knowledge graph entity typing aims to infer entities' missing types in knowledge graphs which is an important but under-explored issue. This paper proposes a novel method for this task by utilizing entities' contextual information. Specifically, we design two inference mechanisms: i) N2T: independently use each neighbor of an entity to infer its type; ii) Agg2T: aggregate the neighbors of an entity to infer its type. Those mechanisms will produce multiple inference results, and an exponentially weighted pooling method is used to generate the final inference result. Furthermore, we propose a novel loss function to alleviate the false-negative problem during training. Experiments on two real-world KGs demonstrate the effectiveness of our method. The source code and data of this paper can be obtained from \url{https://github.com/CCIIPLab/CET}.
\end{abstract}

\section{Introduction}

\begin{figure}[t]
    \centering
    \includegraphics[width=1.\columnwidth]{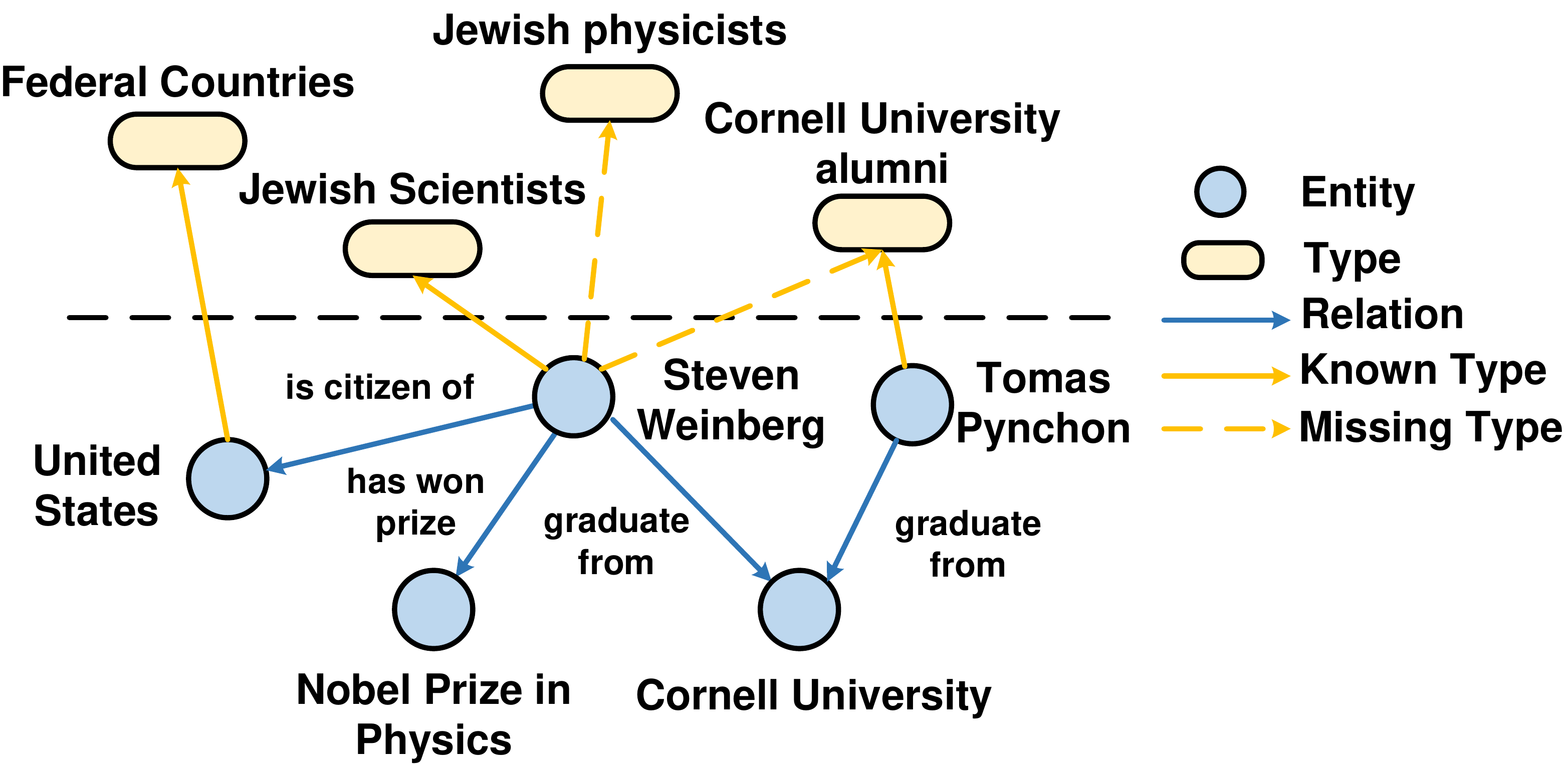}
    \caption{A knowledge graph fragment. Some types of entity \emph{Steven Weinberg} are missing.}
\label{fig:KGexample}
\end{figure}

Knowledge graphs (KGs) store world knowledge in a structured way. They consist of collections of triples in the form of (head entity, relation, tail entity), and entities are labeled with types (see Figure ~\ref{fig:KGexample}). The entity type information on knowledge graph has applications in many NLP tasks including entity linking \citep{gupta2017entity}, question answering \citep{DBLP:conf/emnlp/BordesCW14} and fine-grained entity typing in text  (\citealp{ling2012fine}, \citealp{UFET}, \citealp{zero-shotet}). An entity can have multiple types, and the entity type information on the knowledge graph is usually incomplete.
In this paper, we focus on \emph{Knowledge Graph Entity Typing} (KGET), which aims to infer entities' missing types in knowledge graphs.

Existing methods for the KGET task can be divided into embedding-based methods and graph convolutional networks (GCNs) for the multi-relational graph. Knowledge graph embedding (KGE) is representative of embedding-based methods. Treating entities' known types as special triples with a unique relation "has type", \eg (Barack Obama, has type, person), the KGET task can be understood as a subtask of knowledge graph completion. Consequently, KGE methods can infer entities' missing types by completing (entity, has type, ?). Recently, two embedding-based KGET models based on KGE have been proposed: ETE~\citep{moon2017learning} and ConnectE~\citep{DBLP:conf/acl/ZhaoZXLW20}. They first obtain entity embeddings from KGE methods, then use them to infer entities' missing types. GCNs for the multi-relational graph can aggregate the rich information in entities' neighbors to infer entities' missing types.

Existing methods usually encode all attributes of an entity into one embedding, then use this representation to conduct inference. However, when judging whether an entity has a particular type, only some attributes of this entity may be helpful while the others remain useless. For example, in Figure~\ref{fig:KGexample}, only the neighbor (graduate from, Cornell University) can indicate the central entity \emph{Steven Weinberg} have type \emph{Cornell University alumni}. We argue that always considering all attributes of an entity during inference may introduce irrelevant information as noise and ultimately reduce the accuracy of entity typing.

Besides the above-mentioned shortcoming, ETE and ConnectE also ignore entities' known type information when training entity embeddings, which is important for entity typing. For instance, in Figure \ref{fig:KGexample} there are no known triples which can indicate entity \emph{Steven Weinberg} has type \emph{Jewish physicists}. In this case, the model needs to utilize the known type information. \emph{Steven Weinberg} has type \emph{Jewish Scientists}, and in the known triples exists (Steven Weinberg, has won prize, Nobel Prize in Physics). Combining the two, we can infer \emph{Steven Weinberg} has type \emph{Jewish physicists}. 
In short, these two KGET models have difficult using the entities' known types to infer the missing ones. In the experiment, we found this seriously affect their performance.

To overcome those shortcomings in existing methods, we propose a novel method for the KGET task, called CET (Context-aware Entity Typing). Specifically, CET contains two inference mechanisms: i) N2T: independently use each neighbor of an entity to infer its type; ii) Agg2T: aggregate the neighbors of an entity to infer its type. According to our observation, one neighbor usually represents a specific attribute of the central entity. Thus, the N2T mechanism allows CET to consider each attribute of an entity during inference individually. In contrast, previous works mix various attributes of an entity into one embedding for inference. Therefore, we believe CET can produce more accurate entity typing results than existing methods.
Moreover, some complex types like \emph{21st-century American novelists} involve multiple semantic aspects of an entity. It's difficult to infer those types only using a single neighbor. Therefore, we further propose the Agg2T mechanism, which simultaneously considers multiple attributes of the central entity during inference by aggregating neighbors.
We also treat the known types of the central entity as its neighbors to use them to infer the missing types.
To aggregate the inference results generated by N2T and Agg2T mechanism, we adopt a carefully designed pooling method similar to softpool \citep{DBLP:journals/corr/abs-2101-00440}. Experiments show that this pooling method can produce stable and interpretable inference results.

In addition, we face serious false-negative problem during training. Some (entity, type) pairs are valid but happen to be missing in current knowledge graphs. Treating them as negative samples will seriously affect model performance. We propose a novel loss function to alleviate this.
To sum up, our contributions are three-fold:
\begin{itemize}
    \item We propose CET, a novel and flexible method for inferring entities' missing types in knowledge graphs, which fully utilize the neighbor information in an independent-based mechanism and aggregated-based mechanism.
    \item We design a novel loss function to alleviate the false-negative problem during training.
    \item Experiments on two real-world knowledge graphs demonstrate the superiority of our proposed method over other state-of-the-art algorithms, and the inference process of our method is interpretable.
\end{itemize}

\section{Related Work}

\paratitle{Embedding-based methods.} \citet{moon2017learning} propose to learn type embedding for knowledge graph entity typing and build two methodologies: 
i) Synchronous training: Adding entities' known types to knowledge graphs in the form of triples with a unique relation "has type", \eg (Barack Obama, has type, person), Knowledge Graph Embedding (KGE) methods (\citealp{RESCAL}, \citealp{bordes2013translating}, \citealp{HOLE}) can learn the embeddings of entities and types simultaneously. KGE methods can infer entities' missing types by completing (entity, has type, ?). 
ii) Asynchronous training: The model first obtains entities' embeddings from KGE methods, then minimizes the L1 distance between the entities' and their corresponding types' embeddings while keeping the entities' embeddings fixed. During inference, the smaller L1 distance between an entity and a specific type means the entity is more likely to have this type.
\citet{moon2017learning} observe that there will be only one type of relation associate with types in synchronous training. They claim this lack of diversity of relations means that synchronous training methods have difficulty solving the KGET task. 
So they proposed a model called ETE, which follows the asynchronous training strategy and uses CONTE \citep{conte} to obtain entity embeddings.

\citet{DBLP:conf/acl/ZhaoZXLW20} propose ConnectE, a more advanced KGET model which contains two inference mechanisms. 
One is called E2T, which uses a linear transformation to project the entities' embeddings into type embedding space. 
Another is called TRT, which uses the neighbors' types to infer the central entities' missing types. TRT is based on the assumption that the relationship can remain unchanged when replacing the entities in the triple to their corresponding types. For instance, if triple (Barack Obama, born in, Honolulu) holds, a new triple (person, born in, location) should also hold. 
ConnectE also follows the asynchronous training strategy, which first uses TransE to obtain entities' embedding then fixes them to train E2T and TRT.

ETE and ConnectE do not consider entities' known types when training entities' embeddings, which means they do not encode the known type information into entities' embeddings. Therefore, both of them have difficulty using entities' known types to infer the missing ones, which seriously affects their performance. Our experiments support this claim.

\paratitle{GCNs for Multi-Relational Graph.} The TRT mechanism in ConnectE attempts to use entities' neighbors to infer entities' missing types. However, TRT only utilizes the neighbors' types. To fully utilize the information in entities' neighbors, GCNs for multi-relational graphs can be used to encode entities' neighbors. 
\citet{schlichtkrull2018modeling} proposed R-GCN, an extension of GCNs for relational graphs. R-GCN aggregate the information in neighbors using the relation-specific filter. Weighted Graph Convolutional Network \cite{DBLP:conf/aaai/ShangTHBHZ19} utilizes learnable relational specific scalar weights to aggregate neighbors. \citet{DBLP:conf/iclr/VashishthSNT20} proposed a more generalized framework by leverage composition operators from KGE techniques during GCN aggregation.
In the KGET task, the entities' missing types can be inferred by performing multi-label classification on entities' embeddings obtained by GCNs.

Existing methods usually encode all attributes of an entity into one embedding during inference. We argue this will introduce noise as sometimes only part of attributes of an entity is helpful for the KGET task while the others may be useless. 
To overcome this shortcoming, we propose the N2T mechanism. By independently uses each neighbor of an entity to infer its missing types, the N2T mechanism allows our model to consider each attribute of an entity during inference individually. This can reduce the impact of irrelevant information on entity typing.
Also, we treat entities' known types as neighbors which means our model can use them to infer entities' missing types.

\paratitle{Others.}
Note that embedding knowledge graphs containing concepts (ontologies) and modeling the relationship between concepts (\citealp{Transc}, \citealp{JOIE}) are not the goal of this paper. We concentrate on inferring entities' missing types.
Some other works on KGET (\citealp{neelakantan2015inferring}, \citealp{jin2018attributed}) mainly focus on how to combine additional information, such as the text description of the entities, to infer the missing types. Our work only uses the information on the knowledge graphs to infer the missing types of entities, which is more universal.

\begin{figure*}[th]
    \centering
    \includegraphics[width=1.\textwidth]{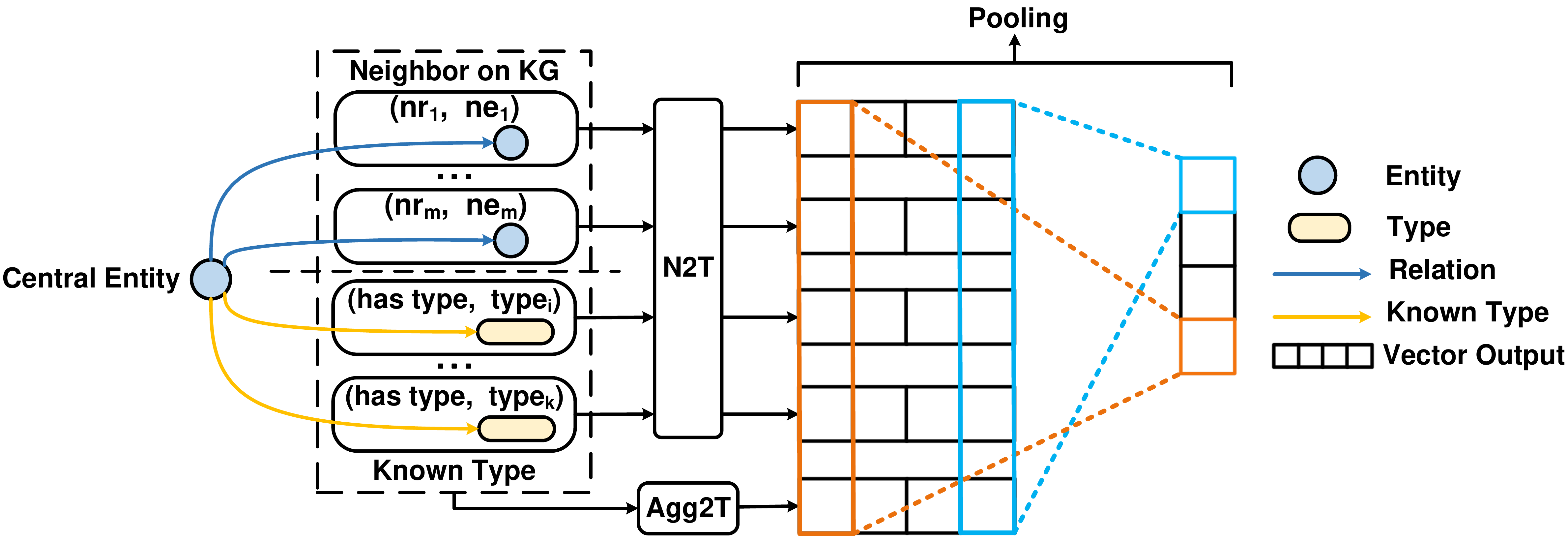}
    \caption{The overall architecture of CET. The N2T mechanism independently uses each neighbor to infer entities' missing types. The Agg2T mechanism aggregates neighbors' information then conducts inference. The final inference result is generated by an exponentially weighted pooling method.}
   \label{fig:model}
\end{figure*}

\section{Method}
In this section, we introduce our proposed method in detail. We first introduce the notations used in this paper. Afterward, we introduce two inference mechanisms used in our method. Finally, we introduce a novel loss function that can alleviate the false negative problem during training.

\subsection{Notations} \label{3.1}
Let $\mathcal{G}=\{(s, r, o)\} \subseteq \mathcal{E} \times \mathcal{R} \times \mathcal{E}$ be a knowledge graph where $\mathcal{E}$ and $\mathcal{R}$ are the entity set and the relation set, respectively. The known type information on the knowledge graph is represented as $\mathcal{I}=\{(e, t)\} \subseteq \mathcal{E} \times \mathcal{T}$. Let $L$ be the number of types. We number the type from 1 to $L$ and use type i to refer to the i-th type.

We add the known type information to the knowledge graphs. If an entity $e$ has type $t$, we add an edge \emph{(e, has type, t)} to KG, where \emph{has type} is a newly added relationship. For the convenience of discussion, if edge $(s, r, o)$ exists in KG, we add its inverted edge $(o,r^{-1},s)$ to KG where $r^{-1}$ is the reverse relation of $r$. Let $\mathcal{G}^{'}$ be the KG after adding known type information and inverted edges. After adding those inverted edges, we only consider outgoing edges when discussing entities' neighbors. The neighbors set of $u$ can be represented as $\mathcal{N}(u)=\{(n_r,n_e)\vert (u, n_r,n_e)\in\mathcal{G}^{'}\}$. We use bold $\vec{n}_r, \vec{n}_e$ to represent the embedding of neighbor relation and neighbor entity, respectively. Let $k$ be the dimension of the embeddings. 
The neighbors mentioned later all refer to those neighbors on $\mathcal{G}^{'}$ which include the neighbors in the knowledge graph and the entities' know types.

\subsection{Proposed Method} \label{3.2}
Our proposed method contains two inference mechanisms. One is to use each neighbor to infer the central entity's type independently, called N2T. Another is to aggregate neighbor information then conduct inference, called Agg2T. And we use an exponentially weighted pooling method to generate the final inference result. The overall architecture is shown in Figure \ref{fig:model}.

\paratitle{N2T mechanism.} We observe a strong correlation between the neighbors and the central entity's type. For instance, the neighbor (is affiliated to, Los Angeles Lakers) can indicate the central entity has type \emph{Los Angeles Lakers player}. Meanwhile, different neighbors may correspond to different types. Therefore, we propose the N2T mechanism that independently uses each neighbor to infer the missing types of central entities. 
It's worth noting that when judging whether an entity has a particular type, sometimes only a few neighbors are helpful while the others remain useless. The N2T mechanism focuses on a single neighbor during inference, reducing the interference of irrelevant information on entity typing.
In practice, CET follows the translating assumption in TransE to obtain the neighbor embedding\footnote{The original relation $r$ and its reversed relation $r^{-1}$ share the same set of parameters and their embeddings satisfy $\vec{r}=-\vec{r}^{-1}$.}, then conducts non-linear activation\footnote{Non-linear activation is not necessary, but we found that adding it can achieve better results.} on neighbor embedding and sent it to a linear layer:
\begin{equation}
    \vec{R}_{(n_r, n_e)}^{N2T} = \matrix{W}\mathrm{Relu}(\vec{n}_e - \vec{n}_r) + \vec{b},
\end{equation}
where $\matrix{W}\in\mathbb{R}^{L\times k}, \vec{b}\in\mathbb{R}^{L}$ are the learning parameters and $\vec{R}_{(n_r, n_e)}^{N2T} \in \mathbb{R}^{L}$ is the relevance score calculated by the N2T mechanism, where the i-th entry represents the relevance score between neighbor $(n_e, n_r)$ and type i. The higher $R_{(n_r, n_e), i}^{N2T}$ means the neighbor $(n_e, n_r)$ is more relevant to type i, which indicates the central entity is more likely to have type i.

\paratitle{Agg2T mechanism.} It's difficult to infer some complex types like \emph{21st-century American novelists} and \emph{Film directors from New York City} from a single neighbor. Therefore, we further propose the Agg2T mechanism which aggregate entities' neighbors to infer entities' missing types:
\begin{equation}
    \vec{h}_u =\frac{1}{\vert\mathcal{N}(u)\vert}\sum_{(n_r, n_e)\in\mathcal{N}(u)}(\vec{n}_e - \vec{n}_r),
\end{equation}
\begin{equation}
    \vec{R}_{u}^{Agg2T} =\matrix{W}\mathrm{Relu}(\vec{h}_u)+\vec{b},
\end{equation}
where $\vec{h}_u\in\mathbb{R}^{k}$ is the aggregated representation of u's neighbors and $\vec{R}_{u}^{Agg2T}\in\mathbb{R}^{L}$ stores the relevance scores with all types. Here we chose a simple non-parameterized mean aggregation operation to verify the effectiveness of our method. Actually, CET is a highly flexible method that can work with the existing GCN-based method by replacing the aggregation operation in the Agg2T mechanism. We leave those analyses as future work.

\paratitle{Pooling approach.} The N2T mechanism and the Agg2T mechanism will generate multiple entity typing results for every entity. To generate the final entity typing result, a pooling method is needed. Mean-pooling is not recommended as some types can only be indicated by a few neighbors. Max-pooling seems to be a suitable choice. However, we find its performance is not ideal. Choosing the max value as the final result makes only a small part of the input get the gradient which means some embeddings may not be sufficiently trained. As a result, the model may fail to represent every attribute of an entity accurately. In practice, we adopt an exponentially weighted pooling method similar to softpool~\citep{DBLP:journals/corr/abs-2101-00440}:
\begin{align}
    & R_{u, i}=\mathrm{pool}(\{R_{u, i}^{Agg2T}, R_{(n_r, n_e), i}^{N2T} \notag \\ 
    & \vert \; \forall (n_e, n_r)\in \mathcal{N}(u)\}),for\; i \in 1,2,\dots, L,
\end{align}
\begin{equation}
    \mathrm{pool}(\{x_1, x_2,...,x_n\})=\sum_{i=1}^{n}w_ix_i,
\end{equation}
\begin{equation} \label{equation:weight}
    w_i = \frac{\exp \alpha x_i}{\sum_{k=1}^{n}\exp \alpha x_k},
\end{equation}
where $R_{u, i}\in \mathbb{R}$ is the relevance score between entity $u$ and type i. $\alpha \in \mathbb{R}^+$ is a hyperparameter that controls the temperature of the pooling process. The higher $R_{u, i}$ means entity $u$ is more likely to have type i. This pooling method has a similar effect to max-pooling but can generate a gradient for every input which ensures every embedding gets sufficient training.

\paratitle{Neighbor sampling.} If we use all the neighbors during training, the model may learn to use available type information to infer themselves, \eg using neighbor (has type, person) to infer the entity has type \emph{person}. The model can perfectly fit the training set in this way, result in a severe overfitting problem. One solution is to perform the following mask operation before the equation (\ref{equation:weight}):
\begin{equation}
    \left \{ 
    \begin{array}{ll}
         R_{(has\;type, i), i}^{N2T} = -\infty, & i\in1,2,\dots,L \\
         R_{u,i}^{Agg2T} = -\infty, & if\; (u, i) \in \mathcal{I}
    \end{array}
    \right.
\end{equation}
Another solution is to perform neighbor sampling: dynamically sample entities' neighbors with replacement during training. We find both methods have similar performance while performing neighbor sampling can significantly save training time, so we finally settle with neighbor sampling. Sampling fewer neighbors can lead to faster training speed, but at the expense of performance degradation. In practice, we find that sampling ten neighbors can usually achieve a good balance between speed and performance. We only conduct neighbor sampling during training; all neighbors of the entity are used during inference.

\subsection{Optimization}
We hope that $R_{u, i}$ as high as possible if entity u has type i (positive samples), while $R_{u, i}$ as low as possible if entity u does not has type i (negative samples). The known (entity, type) pairs in $\mathcal{I}$ can be used as positive samples. To gather the negative samples, a simple choice is to treat all the nonexistent (entity, type) pairs in $\mathcal{I}$ as negative samples. Then we can use the binary cross-entropy (BCE) as loss function:
\begin{equation}
    p_{u, i}=\sigma(R_{u, i}),
\end{equation}
\begin{equation}
    \mathcal{L}=-\sum_{(u, i)\in\mathcal{I}}\log p_{u, i}-\sum_{(u, i)\notin\mathcal{I}}\log(1-p_{u, i}).
\end{equation}
However, some (entity, type) pairs are valid but happen to be missing in current knowledge graphs. Actually, the entities' missing types which we want to infer belongs to this category. This brings serious false negative problems during training. To overcome this, we propose the following false-negative aware (FNA) loss function:
\begin{align}
    \mathcal{L} = & -\sum_{(u, i)\notin\mathcal{I}}\beta(p_{u, i}-p_{u, i}^2)\log(1-p_{u, i}) \notag \\
    & -\sum_{(u, i)\in\mathcal{I}}\log p_{u, i},
\end{align}
where $\beta$ is a hyper-parameter used to control the overall weight of negative samples. The FNA loss function will assign lower weight to those negative samples with too large or too small relevance scores. Those negative samples with too large relevance scores are possibly false negative samples, and those with too small relevance scores are easy ones. These two kinds of negative samples do not provide helpful information, so we give them a lower weight. 

\begin{table}[htb]
  \centering
  \begin{tabular}{lcc}
    \hline
    Dataset & FB15kET & YAGO43kET \\
    \hline
    \# Entities & 14951 & 42334 \\
    \# Relations & 1345  & 37 \\
    \# Types & 3584  & 45182 \\
    \# Train. triples & 483142 & 331686 \\
    \# Train. tuples & 136618 & 375853 \\
    \# Valid & 15848 & 43111 \\
    \# Test & 15847 & 43119 \\
    \hline
  \end{tabular}
  \caption{Statistics of used datasets.}
  \label{tab:datasets}%
\end{table}%

\section{Experiment}
In this section, we evaluate and analyze the proposed method on two real-world KGs. We introduce datasets and experiment settings in Section \ref{4.1}, present the main result in Section \ref{4.2}.  The ablation study can be found in Section \ref{4.3}. Section \ref{4.4} provides some cases to further analyze our method.

\subsection{Datasets and Experiment Setup} \label{4.1}
\paratitle{Datasets.} We conduct experiments on two real-world knowledge graphs, \ie FB15k \citep{bordes2013translating}, YAGO43k \citep{moon2017learning} which are subsets of Freebase \citep{freebase} and YAGO \citep{YAGO}, respectively. \citeauthor{moon2017learning} collected entities' types in both datasets and added them into the original datasets in the form of \emph{(entity, entity type)}. The datasets after adding those entity-type tuples are called FB15kET and YAGO43kET. Their training sets consist of original triples in FB15k and YAGO43k with some entity-type tuples, and the other entity-type tuples are served as validation and test sets. The statistics of the datasets are shown in Table \ref{tab:datasets} \footnote{we exclude the data which contains unseen types in the training set from validation set and test set.}.

\paratitle{Hyper-parameter Settings.} We perform stochastic minibatch training and use Adam \citep{kingma2014adam} as the optimizer. The hyper-parameters are tuned according to the MRR on the validation set. The search space for the grid search are set as follows: embedding dim $k \in \{50, 100\}$, pooling temperature $\alpha \in \{0.5, 1.0\}$, negative samples weight $\beta \in \{1.0, 2.0, 4.0\}$ and learning rate $lr\in\{0.001, 0.005, 0.01\}$. We also tried to adjust the batch size but this had no impact so we fixed the batch size to 128. The embeddings of entities, relations, and types are uniformly initialized, using a uniform distribution:$[-10/k, 10/k]$ (k is the dimension of embeddings). The best model was selected by early stopping using the MRR on validation sets, computed every 25 epochs with a maximum of 1000 epochs. The optima configurations are: $\{k=100, \alpha=0.5, \beta=4.0, lr=0.001\}$ on FB15kET; $\{k=100, \alpha=0.5, \beta=2.0, lr=0.001\}$ on YAGO43kET.

\paratitle{Evaluation Protocol.} For each test sample (\emph{e}, \emph{t}) in test set. We first calculate the relevance score between \emph{e} and every type and then rank all the types in descending order of relevance score. Following \citep{bordes2013translating},  we evaluate model performance in the filtered setting: All the known types of \emph{e} in the training, validation, and test sets are removed from the ranking. Finally, we can obtain the exact rank of the correct type \emph{t} in all types. We use Mean Rank (MR), Mean Reciprocal Rank (MRR), and Hits at 1/3/10 as evaluation metrics.

\paratitle{Baselines.} We compare our model with six state-of-the-art models, which can be divided into three groups. Models in the first group are KGE models which treat the KGET task as a special sub-task of knowledge graph completion, including TarnsE \citep{bordes2013translating}, ComplEx \citep{trouillon2016complex} and RotatE \citep{sun2018rotate}. The second group are recently proposed KGET models including ETE \citep{moon2017learning} and ConnectE \citep{DBLP:conf/acl/ZhaoZXLW20}. And for GCNs for multi-relational graph we choose R-GCN \citep{schlichtkrull2018modeling} as baseline. To make a fair comparison, R-GCN has similar experiment settings with CET: treating entities' known types as neighbors and performing the neighbor sampling during training. Hyper-parameter settings of those baselines can be found in Appendix~\ref{sec:appendix}.

\paratitle{Implementation.} All the KGE baselines in this paper are implemented using DGL-KE \citep{DGL-KE}. Our model and R-GCN are implemented using the DGL framework (PyTorch as backends). All the experiments were run on a single 1080Ti system with 32GB RAM. 

\begin{table*}[htbp]
  \centering
  \resizebox{\textwidth}{!}{
    \begin{tabular}{l|ccccc|ccccc}
    \toprule
    \multicolumn{1}{c|}{\multirow{2}[4]{*}{Model}} & \multicolumn{5}{c|}{\textbf{FB15kET}} & \multicolumn{5}{c}{\textbf{YAGO43kET}} \\
\cmidrule{2-11}          & \textbf{MRR} & \textbf{MR} & \textbf{Hit@1} & \textbf{Hit@3} & \textbf{Hit@10} & \textbf{MRR} & \textbf{MR} & \textbf{Hit@1} & \textbf{Hit@3} & \textbf{Hit@10} \\
    \midrule
    TransE & 0.618  & \textbf{18} & 0.504  & 0.686  & 0.835  & 0.427  & 393   & 0.304  & 0.497  & 0.663  \\
    ComplEx & 0.595  & 20    & 0.463  & 0.680  & 0.841  & 0.435  & 631   & 0.316  & 0.504  & 0.658  \\
    RotatE & 0.632  & \textbf{18}    & 0.523  & 0.699  & 0.840  & 0.462  & 316   & 0.339  & 0.537  & 0.695  \\
    ETE*  & 0.500  & -     & 0.385  & 0.553  & 0.719  & 0.230  & -     & 0.137  & 0.263  & 0.422  \\
    ConnectE*   & 0.590  & -     & 0.496  & 0.643  & 0.799  & 0.280  & -     & 0.160  & 0.309  & 0.479  \\
    R-GCN (BCE) & 0.662  & 19    & 0.571  & 0.711  & 0.836  & 0.357  & 366   & 0.266  & 0.392  & 0.533  \\
    R-GCN (FNA) & 0.679  & 20    & 0.597  & 0.722  & 0.843  & 0.372  & 397   & 0.281  & 0.409  & 0.549  \\
    \midrule
    CET (BCE) & 0.682  & 19    & 0.593  & 0.733  & 0.852  & 0.472  & \textbf{239} & 0.362  & 0.540  & 0.669 \\
    CET (FNA) & \textbf{0.697} & 19    & \textbf{0.613} & \textbf{0.745} & \textbf{0.856} & \textbf{0.503} & 250   & \textbf{0.398} & \textbf{0.567} & \textbf{0.696} \\
    \bottomrule
    \end{tabular}%
  }
  \caption{Results of several models on FB15kET and YAGO43kET datasets. Best results are in \textbf{bold}. [*]: Results are taken from original papers. ConnectE has three different training settings, here we report the best one. }
  \label{tab:mainresult}%
\end{table*}%

\begin{table*}[htbp]
  \centering
  \resizebox{\textwidth}{!}{
    \begin{tabular}{cccc|ccccc|ccccc}
    \toprule
    \multicolumn{4}{c|}{\textbf{Model}} & \multicolumn{5}{c|}{\textbf{FB15kET}} & \multicolumn{5}{c}{\textbf{YAGO43kET}} \\
    \midrule
    \textbf{N2T} & \textbf{TAN} & \textbf{Agg2T} & \textbf{FNA} & \textbf{MRR} & \textbf{MR} & \textbf{Hit@1} & \textbf{Hit@3} & \textbf{Hit@10} & \textbf{MRR} & \textbf{MR} & \textbf{Hit@1} & \textbf{Hit@3} & \textbf{Hit@10} \\
    \midrule
    \checkmark & \checkmark & \checkmark & \checkmark & \textbf{0.697} & \textbf{19} & \textbf{0.613} & \textbf{0.745} & \textbf{0.856} & \textbf{0.503} & 250 & \textbf{0.398} & \textbf{0.567} & \textbf{0.696} \\
    \checkmark & \checkmark & \checkmark &       & 0.682  & \textbf{19} & 0.593  & 0.733  & 0.852  & 0.472  & \textbf{239}   & 0.362  & 0.540  & 0.669  \\
    \checkmark & \checkmark &       &       & 0.679  & \textbf{19} & 0.591  & 0.730  & 0.850  & 0.460  & 272   & 0.348  & 0.528  & 0.664  \\
    \checkmark &       &       &       & 0.663  & 21    & 0.575  & 0.710  & 0.836  & 0.431  & 505   & 0.331  & 0.491  & 0.615  \\
    \bottomrule
    \end{tabular}%
  }
  \caption{Results of ablation study. Models without FNA loss function use BCE loss function instead.}
  \label{tab:ablation study}%
\end{table*}%

\subsection{Main Results} \label{4.2}

Table \ref{tab:mainresult} summarizes our result on FB15kET and YAGO43kET. We implement TransE, ComplEx, and RotatE using the self-adversarial negative sampling \citep{sun2018rotate} and L3 regularization \citep{lacroix2018canonical}, which leads to better results than the reported results in the previous paper. 
We can see our model outperforms all baselines on almost all metrics. Meanwhile, after using the FNA loss, the performance significantly improved. Not only our model, but R-GCN can also benefit from this, which further proved the effectiveness of FNA loss.

The performance of TransE, ComplEx, and RotatE is limited by their entity representation strategy. These KGE methods encode all attributes of an entity into one embedding for inference. However, when judging whether an entity has a particular type, the irrelevant attributes may interfere with the inference result. CET overcomes this shortcoming by using the N2T mechanism and achieves better performance.

Similar to the KGE methods, R-GCN also suffers from the noise introduced by irrelevant information. R-GCN aggregates entities' neighbors to infer entities' missing types. However, sometimes a type can only be indicated by a few neighbors. This kind of rare information is easily overwhelmed by other irrelevant information during R-GCN's aggregation process. This phenomenon is rarely observed on FB15kET but is common on YAGO43kET. This can explain why R-GCN outperforms other baselines on FB15kET but has a poor performance on YAGO43kET.

ETE and ConnectE are largely left behind, especially on YAGO43kET. This is because these two methods have difficulty using entities' known types to infer the missing ones.
Compared with FB15k, YAGO43k has a sparser graph structure and fewer types of relations (see Table \ref{tab:datasets}). Therefore, in YAGO43kET, ignoring entities' known types and only using the (entity, relation, entity) triples to train entity embeddings can hardly fully model various attributes of each entity. As a result, the performance gap between ETE/ConnectE and other methods is more pronounced on YAGO43kET.
This result demonstrates that using entities' known types to infer the missing ones is crucial in the KGET task. We will further illustrate this in Section \ref{4.3}.

\subsection{Ablation Study} \label{4.3}
\begin{table}[htbp]
  \centering
  \resizebox{\columnwidth}{!}{
    \begin{tabular}{c|ccccc}
    \toprule
    \textbf{Model} & \textbf{MRR} & \textbf{MR} & \textbf{Hit@1} & \textbf{Hit@3} & \textbf{Hit@10} \\
    \midrule
    mean  & 0.396  & 338   & 0.300  & 0.440  & 0.578  \\
    max   & 0.462  & 327   & 0.366  & 0.517  & 0.636  \\
    ewp   & \textbf{0.503} & \textbf{250} & \textbf{0.398} & \textbf{0.567} & \textbf{0.696} \\
    \bottomrule
    \end{tabular}%
  }
  \caption{Comparison of different pooling methods.}
  \label{tab:pool}%
\end{table}%

\begin{table*}[htb]
  \centering
    \begin{tabular}{cc|cc}
    \toprule
    \multicolumn{2}{c|}{\textbf{Inference}} & \multicolumn{2}{c}{\textbf{Top 3 Relevant Information Source}} \\
    \midrule
    \textbf{Entity} & \textbf{Type} & \multicolumn{1}{c}{\textbf{Information Source}} & \textbf{Relevance Score} \\
    \midrule
    \multicolumn{1}{c}{\multirow{3}[1]{*}{Bob Dylan}} & \multicolumn{1}{c|}{\multirow{3}[1]{*}{Pulitzer Prize winners}} & (has won prize, Pulitzer Prize) & 6.93 \\
          &       & (has won prize, Quill Award) & -0.44 \\
          &       & (has type, American poets) & -0.72 \\
    \midrule
    \multicolumn{1}{c}{\multirow{3}[0]{*}{Ian Fleming}} & \multicolumn{1}{c|}{\multirow{3}[0]{*}{English writers}} & (has type, English short story writers) & 3.36 \\
          &       & (has type, English novelists) & 2.96 \\
          &       & (has type, English spy fiction writers) & 2.15 \\
    \midrule
    \multicolumn{1}{c}{\multirow{3}[1]{*}{Ben Gazzara}} & \multicolumn{1}{c|}{\multirow{3}[1]{*}{\tabincell{c}{American male \\ television actors}}} & Aggregation & 3.04 \\
          &       & (has type, American male film actors) & -0.53 \\
          &       & (has type, American television actors) & -0.61 \\
    \bottomrule
    \end{tabular}%
  \caption{Representative entity typing examples. We present the top 3 relevant information sources for entity typing and their relevance scores. \emph{Aggregation} stands for the aggregation of neighbors, the information source used in the Agg2T mechanism.}
  \label{tab:cases}%
\end{table*}%

Our model includes two inference mechanisms: N2T and Agg2T. Treating entities' known types as neighbors (TAN, short for types as neighbors) and the false negative aware loss function (FNA) can also improve the performance. To understand each component's effect on the model, we conduct the ablation study on FB15kET and YAGO43kET datasets. The result is reported in Table \ref{tab:ablation study}. We can see the full model (the first row) outperforms all the ablated models on almost all metrics, illustrating every component's effectiveness in our model.

\paratitle{Impact of N2T.} Only using the N2T mechanism, our model still achieves competitive results against other state-of-the-art baselines. This indicates that independently considering entities' different attributes during inference can reduce the noise and produce accurate entity typing results.

\paratitle{Impact of TAN.} Treating entities' known types as neighbors allows CET utilize entities' known type to infer the missing ones. This strategy is especially effective on datasets containing rich entity-type information such as YAGO43kET.

\paratitle{Impact of Agg2T.} Agg2T mechanism is designed to infer those complex types. Types like \emph{21st-century American novelists} that involve multiple attributes of entities and require joint inference by multiple neighbors almost only appear in YAGO43kET. So it is natural that the Agg2T mechanism has little effect on FB15kET but improves model performance on YAGO43kET.

\paratitle{Impact of FNA.} The false-negative aware loss function can bring significant performance improvement, which proves its effectiveness.

We also compared several pooling methods on YAGO43kET. The result is summarized in Table \ref{tab:pool}. \emph{mean}, \emph{max}, \emph{ewp} stand for mean pooling, maximum pooling and exponential weighted pooling, respectively. We can see that exponentially weighted pooling outperforms other pooling methods, which is consistent with our previous analysis.

\subsection{Case Study} \label{4.4}

In Tabel \ref{tab:cases}, we select three representative inferences made by our model. These examples show how CET used the N2T and Agg2T mechanisms to infer entities' missing types, and the inference process is interpretable. 

In the first example, our model mainly uses the neighbor (\emph{has won prize}, \emph{Pulitzer Prize}) to conduct inference. This is intuitive because the correlation between other neighbors and the candidate type \emph{Pulitzer Prize winners} is indeed not strong. In the second example, our model uses several entities' known types to conduct inference. This inference process is reasonable and can be described in natural language: Ian Fleming is an English short story writer, so he is also an English writer. In the first two examples, the model mainly uses the N2T mechanism. However, in the last example, the type \emph{American male television actors} involves multiple attributes of the entity, which the N2T mechanism cannot infer. Therefore, we can see our model uses the aggregation of neighbors to complete the inference, which is consistent with our previous analysis.

In addition, we provide some N2T examples in Table \ref{tab:N2Tcase} to show the mapping from neighbors to types, and the results are intuitive.
\begin{table}[htb]
  \centering
  \resizebox{\columnwidth}{!}{
    \begin{tabular}{c|cc}
    \toprule
    \multirow{2}[4]{*}{\textbf{Neighbors}} & \multicolumn{2}{c}{\textbf{Top 3 Relevant Types}} \\
\cmidrule{2-3}          & \textbf{Type} & \tabincell{c}{\textbf{Relevance} \\ \textbf{Score}} \\
    \midrule
    \multirow{3}[2]{*}{(plays for, A.C. Milan)} & A.C Milan players & 6.32  \\
          & Serie A footballers & 4.68  \\
          & Living people & 2.71  \\
    \midrule
    \multicolumn{1}{c|}{\multirow{3}[2]{*}{\tabincell{c}{(has won prize, \\Nobel Prize in Chemistry)}}} & Nobel laureates in Chemistry & 3.33  \\
          & scientist & 2.35  \\
          & 20th-century chemists & 1.54  \\
    \midrule
    \multicolumn{1}{c|}{\multirow{3}[2]{*}{(type, American rock singers)}} & American rock singers & 6.37  \\
          & American singers & 3.87  \\
          & rock singers & 1.89  \\
    \bottomrule
    \end{tabular}%
  }
    \caption{Three most relevant types with a particular neighbor.}
  \label{tab:N2Tcase}%
\end{table}%

\section{Conclusion}
This paper describes a novel knowledge graph entity typing method called CET, which utilizes the entities' contextual information to infer entities' missing types. We design two inference mechanisms, one is to independently use each neighbor of an entity to infer its types, another is to aggregate entities' neighbors than conduct inference. In addition, we propose a novel loss function to alleviate the false negative problem during training. Our method is highly flexible, and we are considering introducing advanced graph convolutional network technology into our method.

\section*{Acknowledgments}

We would like to thank all the anonymous reviewers for their insightful and valuable suggestions, which help improve this paper's quality. This work is supported by Cognitive Computing and Intelligent Information Processing (CCIIP) Laboratory, School of Computer Science and Technology, Huazhong University of Science and Technology.


\bibliographystyle{acl_natbib}

\appendix

\section{Hyper-parameter Settings}
\label{sec:appendix}
\paratitle{R-GCN.} We use one layer R-GCN with 100-dimension embeddings in our experiment. The hyper-parameters are tuned according to the MRR on the validation set. The search space for the grid search are set as follows: learning rate $lr \in \{0.001, 0.005, 0.01\}$, activation function $\varphi \in {none, relu, tanh}$, and the weight of negative samples in FNA loss $\beta \in \{1, 2, 3, 4\}$. The input embedding is randomly initialized with a uniform distribution [-0.1, 0.1], and the training batch size is fixed to 128. Using \emph{basis-} or \emph{block-diagonal-} decomposition do not improve results but removing the \emph{self-loop} improve performance. Table \ref{tab:RGCNbest} summarizes the best configuration.

\begin{table}[htbp]
  \centering
    \begin{tabular}{ccccc}
    \toprule
    Dataset & lr & $\beta$  & $\varphi$ & self-loop \\
    \midrule
    FB15kET & 0.001  & 3 & none & FALSE \\
    YAGO43kET & 0.001  & 2 & none & FALSE \\
    \bottomrule
    \end{tabular}%
  \caption{The best configuration for R-GCN.}
  \label{tab:RGCNbest}%
\end{table}%

\paratitle{KGE methods.} For KGE methods, we use 200-dimension embeddings in our experiment. We use random search to tune the hyper-parameters for KGE methods. Table \ref{tab:KGEsearchspace} summarizes the search space. \emph{neg\_num} is the number of negative samples for every positive sample; $ \alpha $ is the temperature in the self-adversarial negative sampling; \emph{lr} is the learning rate; $ \lambda $ is the regularization coefficient in L3 regularization; $ \gamma $ is a fixed margin in \emph{logsigmoid} loss function and it also controls the initialization of the embeddings. We fix the training batch size to 1024.

\begin{table}[htbp]
  \centering
  \resizebox{\columnwidth}{!}{
    \begin{tabular}{c|ccccc}
    \toprule
    Model & neg\_num & $\alpha$ & lr & $\lambda$ & $\gamma$ \\
    \midrule
    TransE & \{128, 256, 512\} & [0.5, 2.0] & [0.005, 0.2] & [1e-7, 1e-5] & [5, 15] \\
    ComplEx & \{128, 256, 512\} & [0.5, 2.0] & [0.005, 0.2] & [1e-7, 1e-5] & [60, 80] \\
    RotatE & \{128, 256, 512\} & [0.5, 2.0] & [0.005, 0.2] & [1e-7, 1e-5] & [5, 20] \\
    \bottomrule
    \end{tabular}%
  }
  \caption{Search space for KGE methods.}
  \label{tab:KGEsearchspace}%
\end{table}%

We run 100 trails for each model, and every trial runs 50000 steps. The best configuration was selected according to the MRR on the validation set. Table \ref{tab:KGEbest} summarizes the best configuration for each model.

\begin{table}[htbp]
  \centering
  \resizebox{\columnwidth}{!}{
    \begin{tabular}{c|c|ccccc}
    \toprule
    Dataset & Model & neg\_num & $\alpha$ & lr & $\lambda$ & $\gamma$ \\    
    \midrule
    \multirow{3}[2]{*}{FB15kET} & TransE & 256   & 1.98  & 0.023 & 7.20E-06 & 6.5 \\
          & ComplEx & 512   & 2.00     & 0.148 & 6.20E-06 & 66.8 \\
          & RotatE & 512   & 1.91  & 0.0168 & 3.50E-06 & 6.0 \\
    \midrule
    \multirow{3}[2]{*}{YAGO43kET} & TransE & 512   & 1.99  & 0.05  & 4.40E-06 & 10.5 \\
          & ComplEx & 512   & 1.99  & 0.154 & 2.00E-06 & 62.4 \\
          & RotatE & 256   & 1.74  & 0.0344 & 2.40E-06 & 11.8 \\
    \bottomrule
    \end{tabular}%
  }
  \caption{The best configuration for KGE methods.}
  \label{tab:KGEbest}%
\end{table}%

\end{document}